\documentclass[letterpaper]{article} 
\usepackage{aaai25}  
\usepackage{times}  
\usepackage{helvet}  
\usepackage{courier}  
\usepackage[hyphens]{url}  
\usepackage{graphicx} 
\usepackage{natbib}  
\usepackage{caption} 
\usepackage{algorithm}
\usepackage{algorithmic}

\usepackage{booktabs} 
\usepackage{multirow}

\usepackage{amsmath}
\usepackage{amssymb}

\usepackage{newfloat}
\usepackage{listings}
\DeclareCaptionStyle{ruled}{labelfont=normalfont,labelsep=colon,strut=off} 
\floatstyle{ruled}
\newfloat{listing}{tb}{lst}{}
\floatname{listing}{Listing}
\title{GLAD: Improving Latent Graph Generative Modeling \\ with Simple Quantization}
\author{
    Van Khoa Nguyen,
    Yoann Boget,
    Frantzeska Lavda,
    Alexandros Kalousis
}
\affiliations{
    Geneva School for Business Administration (HES-SO)\\
    University of Geneva,
    1214 Geneva, Switzerland
    
    \{van-khoa.nguyen, yoann.boget, frantzeska.lavda\}@etu.unige.ch, alexandros.kalousis@unige.ch
}

\begin{document}

\maketitle

\begin{abstract}
Learning graph generative models over latent spaces has received  less attention compared to models that operate on the original data space and has so far demonstrated lacklustre performance. We present GLAD a latent space graph generative model. Unlike most previous latent space graph generative models, GLAD operates on a discrete latent space that preserves to a significant extent the discrete nature of the graph structures making no unnatural assumptions such as latent space continuity. We learn the prior of our discrete latent space by adapting diffusion bridges to its structure. By operating over an appropriately constructed latent space we avoid relying on decompositions that are often used in models that operate in the original data space. We present experiments on a series of graph benchmark datasets that demonstrates GLAD as the first equivariant latent graph generative method achieves competitive performance with the state of the art baselines.
\end{abstract}

\begin{links}
    \link{Code}{https://github.com/v18nguye/GLAD}
\end{links}

\section{Introduction}
Graph generation has posed a longstanding challenge with very diverse methods proposed over time. Most of them operate directly over the original data space and learn generative models over node and edge features
\cite{shi2020graphaf, luo2021graphdf, kong2023autoregressive}. 
Considerably less attention has been given to methods that operate over graph latent spaces and the appropriate definition and design of such latent spaces. 
Some initial efforts \cite{simonovsky2018graphvae, jin2018junction, liu2018constrained, samanta2020nevae} propose learning the graph distribution over the continuous latent space of variational autoecoders (VAEs) \cite{kingma2013auto}. Yet, such approaches often suffer from high reconstruction errors, have to solve challenging graph-matching problems, and/or rely on heuristic corrections for the generated graphs.

Diffusion-based models have recently emerged as a compelling approach for modeling graph distributions \cite{niu2020permutation, jo2022score, vignac2023digress}. They allow for the natural incorporation of permutation-invariance as an inductive bias within their denoising component;  nevertheless, they are not without limitations.
The initial graph diffusion methods rely on continuous score functions to model the graph distributions. They learn these score functions by denoising graph structures that have been noised with real-valued noise
\cite{niu2020permutation, jo2022score}. Continuous score functions are a poor match for structures that are inherently discrete. A rather limited number of diffusion-based works treat graphs heads-on as discrete objects \cite{vignac2023digress}.  
Moreover, constructing appropriate score functions over the original graph representation is challenging. Common approaches factorise the score function to node- and adjacency-matrix-specific components, which require their own distinct denoising networks. These partial score functions provide a fragmented view of the graph and do not reflect well the true score function making them a poor choice to model graph distributions.
Diffusion-based methods typically have a denoising component which operates over the edges. The denoising process takes place over dense, fully-connected, graphs \cite{niu2020permutation, jo2022score, vignac2023digress}, posing significant scalability challenges \cite{qin2023sparse}.

In this work we start from certain desiderata that a graph generative model should satisfy, which will guide the design of our model. Graph generative models should learn graph distributions that are invariant to node permutations. 
Graphs should be treated in a holistic manner without relying on artificial decompositions to their components.
Finally graphs should be treated as discrete objects, as they are.
To address these desiderata we propose \textbf{G}raph Discrete \textbf{La}tent \textbf{D}iffusion (GLAD) a  generative model that operates on a carefully designed permutation-equivariant discrete latent space that explicily accounts for the discrete nature of graphs.
We quantize our latent node embeddings to preserve their discrete nature in the graph space a fact that ensures efficient graph encoding and supports highly accurate reconstructions, compared to existing methods operating on continuous latent spaces. Subsequently, we tailor diffusion bridges, \cite{liu2023learning} proposed for constrained data domains, to model latent graph distributions over the quantized space. Our latent diffusion model relies on a 
geometric set of latent nodes, where latent node coordinates encode local graph substructures, and does not resort to a component specific decomposition as other diffusion models operating on the graph space do. To the best of our knowledge, GLAD is the first equivariant latent graph diffusion model that models in a permutation-invariant manner graph distributions. We carefully evaluate GLAD on a set of graph benchmarks that clearly show that it excels in capturing both chemical and structural properties of graphs compared to state-of-the-art graph generative baselines.

\section{Graph Latent Spaces}

In this section, we discuss existing graph generative methods, how they structure their latent spaces, and their limitations that motivated the design of the latent space of GLAD. 

\paragraph{Continuous-graph latent spaces} \citeauthor{simonovsky2018graphvae} were probably the first to use variational autoencoders (VAEs), \cite{kingma2013auto}, for graph generative modeling. They establish a global latent representation of graph by pooling its node embeddings. The posterior distribution of the graph encoding lies in a continuous space. The use of pooling breaks the permutation symmetry
and requires solving a graph-matching problem in order to compute the graph-reconstruction loss. Graph-matching scales poorly and is inaccurate for large graphs.

\begin{figure*}[t]
\begin{center}
\centerline{
\includegraphics[width=0.75\textwidth]{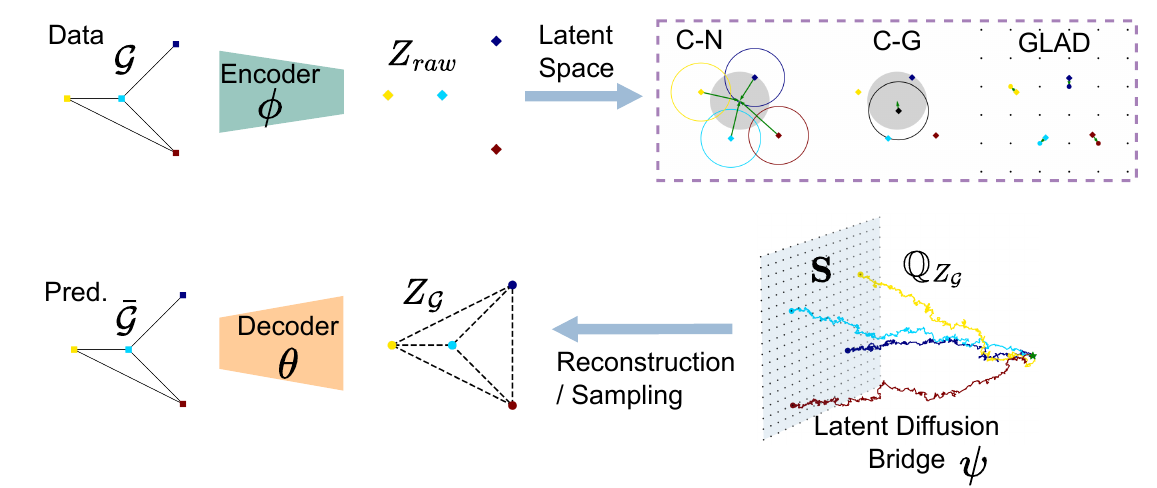}}
\caption{(Top) We encode an annotated graph with node features (colored squares) to a set of node embeddings (colored diamonds). We visualize three studied graph-latent spaces: continuous-graph latent (C-G), continuous-node latent (C-N), and quantised latent (GLAD) spaces. The grey disks denote normal priors, the colored circles denote posterior distributions, and the green arrows denote pushing forces. GLAD distinctively maps the raw-node embeddings to discrete points on a uniformly quantized space (black dots). (Bottom) We decode a fully-connected pseudo latent graph (dashed-edge graph) formed by the set of quantized latent nodes $Z_{\mathcal{G}}$ (colored dots) to the original graph space. We leverage diffusion bridges to learn the graph-latent discrete distribution $\Pi$  constrained over the quantized-latent space $\mathbf{S}$. The colored paths represent an example of $Z_{\mathcal{G}}$-bridge.
}
\label{fig:model-pipe}
\end{center}
\end{figure*}

\paragraph{Continuous-node latent spaces} Subsequent works, \cite{li2018multi, samanta2020nevae}, that follow the VAE approach choose to encode graphs as sets of node embeddings. Operating over node embeddings allows for the use of standard $l_2$ reconstruction loss and removes the need for solving the graph matching problem. Such approaches typically define the posterior graph distribution as the product of the posterior node distributions, with each posterior node distribution being regularised towards the same prior, typically the standard Gaussian prior. As we will empirically demonstrate the result of this regularisation is that the posterior node distributions are indistinguishable between them. We call this indistinguishability latent-node collapse and show that it drastically hinders graph reconstruction. In 
 Figure \ref{fig:model-pipe} we provide a cartoon visualisation of the continuous graph and node latent spaces. 

\paragraph{Discrete latent spaces} Obtaining distinguishable encodings of local graph structures is necessary if one wants to (i) improve the graph reconstruction performance and (ii) ensure that the learnt generative models capture diverse local graph patterns. These objectives can be achieved with an appropriate design of the discrete latent space, where the encodings of the nodes and their neighborhoods (subgraphs) are embedded in a distinguishable manner in the latent space. The main challenge is to preserve the discrete nature of the graphs in a meaningfully manner when we encode them to the latent spaces. There are few notable works that treat latent graph generative modelling as a discrete problem \cite{luo2021graphdf, boget2024discrete}. The former builds an auto-regressive discrete flow over a discrete latent space defined by a modulo shift transform. The latter relies on vector quantization \cite{van2017neural} to encode the latent graph representations using codebooks, and learns the discrete latent graph distribution auto-regressively. While these methods have demonstrated a superior performance over their continuous counterparts \cite{shi2020graphaf}, they still belong to the auto-regressive family and require a canonical-node ordering and as a consequence do not model graph distributions in a permutation-invariant manner.   
In this work, we introduce a novel discrete latent space for graphs, which extends the finite-quantization \cite{mentzer2023finite} to the graph generation setting. Our graph latent space is simple in design, yet effective in performance. Over the graph latent space we learn an equivariant generative model that models graph distributions in a permutation-invariant manner. 

\section{Graph Discrete Latent Diffusion Model}
We will now introduce GLAD, the first graph equivariant generative model that operates on a discrete latent space of graphs. We first present our mathematical notations, then define a novel discrete graph latent space, and show how to learn a prior over the defined latent space using diffusion bridges.

\subsection{Notation}
\label{sec:notation}
We denote a graph by the tuple $\mathcal{G}=(X, E)$; $X = \{ x_i\}_{i=1}^N$ is the set of the node feature vectors, $E = \{ e_{ij} \}_{i,j \in N}$ is the set of edge feature vectors and $N$ is the number of nodes. We rely on an autoencoder (AE) to embed graphs to a latent space. The encoder ($\phi$)-, and decoder ($\theta$)- are equivariant graph neural networks (E-GNNs). 
In the latent space, we denote by $Z_{raw} = \{z_i\}_{i=1}^N$ the set of node embeddings, where $z_i \in \mathbb{R}^{f}$ corresponds to the embedding of the $i^{th}$ graph node. 

\subsection{GLAD's Discrete Graph Latent Space}
\label{sec:discrete}
In constructing the latent space of GLAD we want to satisfy the following desiderata: 
i) ensure that we learn graph distributions that are invariant to permutations, 
ii) encode nodes and their local structures in a way that preserves their structural specificities, and 
iii) preserve the discrete nature of the objects we encode, nodes and their neighborhoods, and that of the graph itself.

We use an E-GNN encoder, $\phi$, to map a graph to the set of the continuous node embeddings $Z_{raw} = \{z_i\}_{i=1}^N := \phi(\mathcal{G})$. The encoder captures node and node-neighborhood information. We then apply the quantization operator, $\mathcal{F}$, on the continuous node embeddings and map them to a quantized space of the same dimension. We follow \cite{mentzer2023finite} to design the quantisation $\mathcal{F}$, which we apply independently on each latent node dimension. The quantization of the $j^{th}$ dimension of the $z_i$ node embedding is given by:
\begin{equation}
    \label{eq:quantize}
    z^q_{ij} = \mathcal{F}(z_{ij}, L_j) = \mathcal{R}(\frac{L_j}{2}\tanh{z_{ij}}),\; 1  \leqslant j  \leqslant f
\end{equation}
Where $\mathcal{R}$ is the rounding operator, and $L_j$ is the number of quantization levels on the $j^{th}$ latent node dimension. 
We will denote by $[L_j]$ the set of pre-defined quantization levels: $[- \mathcal{R}(L_j/2), \cdots, -1, 0, 1, \cdots, \mathcal{R}(L_j/2)]$.
The quantization creates a new set of now discrete latent node embeddings $Z_{\mathcal{G}} = \{z_i^q\}_{i=1}^N$ with $z_i^q \in \mathbf{S}$; $\mathbf{S}$ is the discrete 
latent space which is given by the Cartesian product $ \mathbf{S} := \prod_{j=1}^f [L_j] = \{ l_1 \times \cdots \times l_f:  \forall l_j \in [L_j]\}$, we denote by $K$ the cardinality of $\mathbf{S}$.

The operator $\mathcal{F}$ is permutation equivariant and so is the mapping from the initial graph $\mathcal G$ to its quantized latent representation as long as the $\phi$ 
graph encoder is permutation equivariant. Thus for any $P$ permutation matrix the following equivariance relation holds:
\begin{equation}
    P^T Z_{\mathcal{G}} = \mathcal{F}(P^T Z) = \mathcal{F}(\phi(P^T E P, P^T X)) 
\end{equation}
We apply an equivariant decoder, $\theta(Z_{\mathcal{G}})$,  assuming a fully-connected graph on the quantized latent nodes $Z_{\mathcal{G}}$. 
Our graph autoencoder is composed exclusively of equivariant components. Therefore, we ensure that decoded distributions are invariant 
to the graph node permutations. 
The decoded nodes have a one to one relation to the nodes of the original graph, making it trivial to define a reconstruction loss for the 
autoencoder. We will now show how to use diffusion bridges to learn the discrete latent graph distribution on the latent space induced by
our autoencoder.

\begin{algorithm}[t]
   \caption{Graph Discrete Latent Diffusion Bridge}
   \label{alg:glad}
\begin{algorithmic}
    \STATE {\bfseries Input} Encoder $\phi$, decoder $\theta$, bridge model $\psi$ \\
    
    // \textit{Stage 1: Learn graph autoencoder ($\phi$, $\theta$)}
    
    \STATE  {\bfseries for} batch $\mathcal{G}=(X, E)$:
    \STATE \hspace{5pt} $Z_{raw} \gets \phi(\mathcal{G})$
    \STATE \hspace{5pt} $Z_{\mathcal{G}} \gets \mathcal{F}(Z_{raw}, L)$
    \STATE \hspace{5pt} $Z_{\mathcal{G}} \gets \textit{straight-through-estimator}(Z_{\mathcal{G}}, Z_{raw})$
    \STATE \hspace{5pt} $\mathcal{L}_1 \gets \left\|X - \theta_{X}(Z_{\mathcal{G}})\right\|^2 + \left\|E - \theta_{E}(Z_{\mathcal{G}})\right\|^2$ 
    \STATE \hspace{3pt} $\phi, \theta \gets \textit{Adam-optim} \left(\nabla_{\phi, \theta} \left(\mathcal{L}_1\right)\right)$ \\

    // \textit{Stage 2: Learn bridge model ($\psi$)}
    \STATE {\bfseries for} batch ($Z_\mathcal{G}, t \sim [0, T], Z_0 \sim \mathbb{P}_0)$:
    \STATE \hspace{5pt} $Z_t \gets \frac{\beta_t}{\beta_T}\left(Z_{\mathcal{G}} + (\beta_T -\beta_t)Z_0\right)+ \xi\sqrt{\beta_t\left(1- \frac{\beta_t}{\beta_T}\right)}$
    \STATE \hspace{5pt} $ \left[\eta^{Z_\mathcal{G}}\right]^{I_i} \gets \sigma_t^2\frac{Z_{\mathcal{G}}^{I_i} - Z_t^{I_i}}{\beta_T - \beta_t}$ \hspace{5pt} $\forall I_i \in \Omega$
    \STATE \hspace{5pt} $\left[\eta^{\Pi}\right]^{I_i} \gets \sigma_t^2 \nabla_{Z_t^{I_i}}\log\sum_{s_k \in \mathbf{S}} \exp\left(-\frac{\| Z_t^{I_i} - s_k \|^2}{2(\beta_T - \beta_t)})\right)$
    \STATE \hspace{5pt} $\mathcal{L}_2 \gets \sum_{I_i \in \Omega} \left\| \left[{\psi}\left(Z_t, t\right)\right]^{I_i} - \sigma^{-1}_t\left(\left[\eta^{Z_{\mathcal{G}}}\right]^{I_i}- \left[\eta^{\Pi}\right]^{I_i}\right)\right\|^2$
    \STATE \hspace{5pt} ${\psi} \gets \textit{Adam-optim} \left(\nabla_{\psi} \left(\mathcal{L}_2\right)\right)$\\
    
    // \textit{Sampling novel graphs}
    \STATE $t \gets 0$, $\delta_t \gets \frac{1}{T}$, $Z_0 \sim \mathbb{P}_0$
    \STATE  {\bfseries while} ($t < T$):
    \STATE \hspace{5pt} $\xi \sim \mathcal{N}(0, I) $
    \STATE \hspace{5pt} $ Z_{t+1} \gets Z_{t} + \left(\sigma_t{\psi}(Z_t,t) + \eta^{\Pi}(Z_t,t)\right)\delta_t + \sigma_t\sqrt{\delta_t}\xi$
    \STATE $\mathcal{G}_{new} \gets \theta(Z_T)$
\end{algorithmic}
\end{algorithm}

\subsection{Learning Graph Prior with Diffusion Bridges}
\label{sec:prior}

For a given graph, the quantization maps the set of original node embeddings to points in the discrete latent space $\mathbf{S}$. The latent-graph domain $\Omega$ becomes a set structure that we can decompose into a node-wise manner as $\Omega = I_1 \times \cdots \times I_N$, where $I_i = \mathbf{S}$. We extend diffusion bridges \cite{liu2023learning}, which learn data distributions on constrained domains, to model our latent-graph structure $\Omega$. 
To do so we will first define a conditional diffusion process, conditioned on a given latent graph structure $Z_{\mathcal{G}}$. We then build the diffusion bridge for our latent graph 
distribution  $\Pi$  as the mixture of conditional diffusion processes defined over the training data. 

We start by defining a Brownian motion-based non-conditional diffusion process given by the stochastic differential equation (SDE):
\begin{equation}
    \label{eq:Q}
    \mathbb{Q}: dZ_t = \sigma(Z_t,t)dW_t
\end{equation}
where $W_t$ is a Wiener process and $\sigma: [0,T] \times \mathbb{R}^d \mapsto \mathbb{R}$ is a diffusion coefficient. 

\paragraph{$Z_{\mathcal{G}}$-conditioned diffusion bridge} We will now define a conditional diffusion bridge where the endpoint $Z_T$ of the diffusion process will be the conditioning factor $Z_{\mathcal{G}}$, i.e. $Z_T = Z_{\mathcal{G}}$.
We denote the $Z_{\mathcal{G}}$-conditioned bridge by $\mathbb{Q}^{Z_{\mathcal{G}}}$, the dynamics of which can be derived from the \textit{h-transform} \cite{oksendal2013stochastic}:
\begin{equation}
\label{eq:eta_x}
\mathbb{Q}^{Z_{\mathcal{G}}}:  dZ_t = \eta^{Z_{\mathcal{G}}}(Z_t,t)dt + \sigma(Z_t,t)dW_t, \;\;\; Z_0 \sim \mathbb{P}_0
\end{equation}
where  $\mathbb{P}_0$ is the prior distribution of the $Z_{\mathcal{G}}$-conditione bridge.
The $Z_{\mathcal{G}}$-conditioned bridge's drift is defined as $ \eta^{Z_{\mathcal{G}}}(Z_t,t) = \sigma^2(Z_t,t)\nabla_{Z_t}\log q_{T|t}(Z_{\mathcal{G}}|Z_t)$. $q_{T|t}(Z_{\mathcal{G}}|Z_t)$ is the probability density of obtaining $Z_{\mathcal{G}}$ at time $T$ when we have $Z_t$ at time $t$; 
$\nabla_{Z_t}\log q_{T|t}(Z_{\mathcal{G}}|Z_t)$ 
acts as a steering force, which is central to guiding $Z_t$ towards our specified target, $Z_T=Z_{\mathcal{G}}$.

\paragraph{$\Pi$-conditioned diffusion bridge} Given the $Z_{\mathcal{G}}$-conditioned bridge definition, we can now construct a bridge process on the discrete latent-graph distribution $\Pi$, which is constrained over the latent-graph domain $\Omega$. We call this bridge process a $\Pi$-bridge  and denote it by $\mathbb{Q}^{\Pi}$.
We construct the  $\Pi$-bridge as a mixture of $Z_{\mathcal{G}}$-conditioned bridges; their end-points are conditioned on latent-graph samples $\{Z_{\mathcal{G}}^{i}\}$ i.i.d drawn from the latent graph distribution $\Pi$.
The  $\Pi$-bridge's dynamics are governed by the SDE:
\begin{equation}
\label{eq:eta_o}
\mathbb{Q}^{\Pi}:  dZ_t = \eta^{\Pi}(Z_t,t)dt + \sigma(Z_t,t)dW_t
\end{equation}
The drift  is given by $\eta^{\Pi}(Z_t,t) = \sigma^2(Z_t,t)\mathbb{E}_{\omega \sim q_{T|t,\Omega}(\omega|Z_t)}[\nabla_{Z_t}\log q_{T|t}(\omega|Z_t)]$, where $\omega$ is sampled from a transition probability density given by $q_{T|t, \Omega}(\omega|Z_t) = q(Z_T = \omega| Z_t, Z_T \in \Omega)$.
We see that the discrete latent-graph structure $\Omega$ is taken into consideration in the expectation of the steering forces. The marginal distribution, $\mathbb{Q}_T^{\Pi}$, induced by the $\Pi$-bridge at $t=T$ will match the latent graph distribution $\Pi$ by construction, i.e. $\mathbb{Q}_T^{\Pi} = \Pi$.

\begin{figure}[t]
\begin{center}
\centerline{
\includegraphics[width=1.\columnwidth]{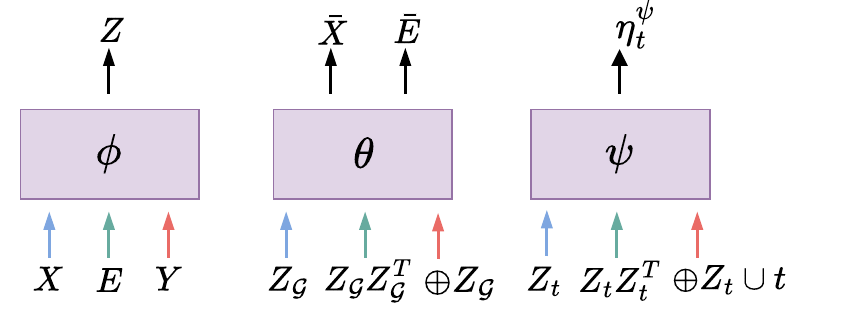}}
\caption{ Graph transformer-based architectures. (Left) graph encoder,  (Middle) graph decoder, (Right) model bridge drift. We denote  $\oplus$ as pooling operator on a set of latent nodes, and $\cup$ as concatenation operator.
}
\label{fig:model-arch}
\end{center}
\end{figure}

\paragraph{Model bridge} We will train a bridge, $\mathbb{P}^{\psi}$, to fit, and generalise from, the $\Pi$-bridge dynamics and generate new latent-graph samples from the $\Pi$ distribution. 
The dynamics of  $\mathbb{P}^{\psi}$ are given by the parametric SDE:
\begin{equation}
    \label{eq:model-bridge}
    \mathbb{P}^{\psi}: dZ_t = \eta^{\psi}(Z_t,t)dt + \sigma(Z_t,t)dW_t
\end{equation}
where the drift term is $\eta^{\psi}(Z_t,t) = \sigma(Z_t, t){\psi}(Z_t,t) + \eta^{\Pi}(Z_t,t)$; ${\psi}$ is a neural network. The ${\Pi}$-bridge imputes noise to latent-graph samples at different time steps which we use to learn the model bridge $\mathbb{P}^{\psi}$. 
We train $\mathbb{P}^{\psi}$ by minimizing the Kullback-Leibler divergence between the two probability path measures $\min_\psi \{ \mathcal{L}(\psi) := \mathcal{KL}(\mathbb{Q}^{\Pi} || \mathbb{P}^{\psi})\}$.
The Girsanov theorem, \cite{lejay2018girsanov}, allows us to compute the $\mathcal{KL}$ divergence with the following closed form solution:
\begin{equation}
    \label{eq:obj}
    \mathcal{L} = \mathbb{E}_{Z_{\mathcal{G}} \sim \Pi, t \sim [0,T], Z_t \sim \mathbb{Q}^{Z_{\mathcal{G}}}} \left[\|{\psi}(Z_t,t) - \Bar{\eta}(Z_t,t)\|^2\right]
\end{equation}
with $\Bar{\eta}(Z_t,t) = \sigma^{-1}(Z_t,t)(\eta^{Z_{\mathcal{G}}}(Z_t,t) - \eta^{\Pi}(Z_t,t))$, where $\eta^{Z_{\mathcal{G}}}$, $\eta^{\Pi}$ are introduced in Equation \ref{eq:eta_x} and Equation \ref{eq:eta_o}, respectively. Since we use Brownian motion for the non-conditional diffusion $\mathbb{Q}$, we can retrieve its closed-form perturbation kernel and accordingly derive the dynamics of the $Z_{\mathcal{G}}$-bridge's drift as:
\begin{equation}
\label{eq:drift_x}
    \eta^{Z_{\mathcal{G}}}(Z_t, t) = \sigma_t^2\frac{Z_{\mathcal{G}} - Z_t}{\beta_T - \beta_t}, \; \beta_t = \int_0^t\sigma_s^2ds
\end{equation}
Where we simply define the diffusion coefficient $\sigma_t$ that depends only on the time variable. As our latent-graph structure  $\Omega$ can be factorized into the latent-structure of nodes, the expectation over $\Omega$ , Equation \ref{eq:eta_o}, can simplify to one-dimensional integrals over $\mathbf{S}$. And the $\Pi$-bridge's drift can be decomposed into the latent-node structures:
\begin{align}
    \eta^{\Pi}(Z_t, t) &= \left[\left[\eta^{\Pi}(Z_t^i, t)\right]^{I_i}\right]_{i=1}^N   \nonumber \\  
    \left[\eta^{\Pi}(Z_t^i, t)\right]^{I_i} &= \sigma_t^2 \nabla_{Z_t^i}\log\sum_{k=1}^K \exp\left(-\frac{\| Z_t^i - s_k \|^2}{2(\beta_T - \beta_t)}\right) \label{eq:drift_pi}
\end{align}
where $s_k \in \mathbf{S}$, $Z_t^i$ is the feature of $i^{th}$ latent node of $Z_t$. We apply the drift terms obtained above to Equation \ref{eq:obj} to get the closed-form objective.

\paragraph{Training}
We train GLAD in two stages. First, we train a graph autoencoder that learns a structural mapping from the graph space to the designed latent space. As a technical detail, we note that the rounding operator $\mathcal{R}$ of the quantization is non-differentiable. To ensure end-to-end training of our discrete graph latent space, we use the straight-through estimator (STE) \cite{bengio2013estimating}, implemented with a stop-gradient operator $\mathcal{S}$ as $Z_{\mathcal{G}} \mapsto Z + \mathcal{S}(Z_{\mathcal{G}} - Z)$. We freeze the graph-autoencoder while training the model bridge in the second stage. GLAD learns the graph latent distributions constrained over the quantized space defined on the graph-encoder output and the quantization operator. At sampling, we apply a simple discretization scheme Euler–Maruyama on the model bridge. We summarise our training procedure in Algorithm \ref{alg:glad}.

\paragraph{Architectures}

We learn data distributions that lie on both graph- and set-based structures in GLAD. As sets are more or less considered as fully-connected graphs. We opt to employ a general-purpose architecture such as graph transformers \cite{dwivedi2020generalization}. Future works could consider a more specific choice of architecture designed to graph- or set-related domains. We adapt the graph transformer \cite{vignac2023digress} and further customize the architecture for our graph encoder $\phi$, graph decoder $\theta$, and drift model ${\psi}$. We visualize our architectures in Figure \ref{fig:model-arch}.
The graph encoder takes a node feature matrix $X$, an edge feature matrix $E$, and a spectral feature vector $Y$ as inputs. Due to the inherent limitations of most message-passing neural networks \cite{chen2020can}, we compute structural features like cycle counts and spectral features of graph-Laplacian decomposition as hand-crafted input features $Y$. The encoder outputs a set of raw node embeddings that is quantized to obtain a set of quantized latent nodes $Z_{\mathcal{G}}$. The graph decoder takes a pseudo-latent graph, which is formed by the latent nodes $Z_{\mathcal{G}}$, a pseudo-adjacency matrix $Z_{\mathcal{G}}Z_{\mathcal{G}}^T$; in addition, we incorporate  a global representation feature to the decoder by pooling the set of latent nodes  $\oplus Z_{\mathcal{G}}$. The drift model is similar in the design of graph decoder, however it takes a noisy set of latent nodes $Z_t$ sampled at different time step $t$ from $Z_{\mathcal{G}}$-bridge. The model learns the dynamic of $\Pi$-bridge drift by $\eta_t^{\psi}$, which structure is identical to $Z_{\mathcal{G}}$.

\begin{table*}[t]
\small
\begin{center}
\begin{tabular}{l cccc cccc cccc}
\toprule
 & \multicolumn{4}{c}{\textsc{Community-small}} & \multicolumn{4}{c}{\textsc{Ego-small}} & \multicolumn{4}{c}{\textsc{Enzymes}}\\
\cmidrule(l){2-5} \cmidrule(l){6-9} \cmidrule(l){10-13}
 & Deg. $\downarrow$ &  Clus. $\downarrow$ & Orb. $\downarrow$ & AVG & Deg. $\downarrow$ &  Clus. $\downarrow$ & Orb. $\downarrow$ & AVG & Deg. $\downarrow$ &  Clus. $\downarrow$ & Orb. $\downarrow$ & AVG \\
\midrule
GraphRNN  & 0.080 & 0.120 & 0.040 & 0.080  & 0.090 & 0.220 & \underline{0.003} & 0.104 & 0.017 & 0.062 & 0.046 & 0.042\\
GraphAF  & 0.180 & 0.200 & 0.020 & 0.133  & 0.030 & 0.110 & \textbf{0.001} & 0.047 & 1.669 & 1.283 & 0.266 & 1.073\\
GDSS  & 0.045 & 0.086 & 0.007 & 0.046  & 0.021 & 0.024 & 0.007 & 0.017 & 0.026 & 0.061 & 0.009 & 0.032 \\
DiGress & 0.047 & \textbf{0.041} & 0.026 & 0.038  & \underline{0.015} & 0.029 & 0.005 & 0.016  & \textbf{0.004} & 0.083 & \underline{0.002} & 0.030 \\
GraphArm  & 0.034 & 0.082 & \textbf{0.004} & 0.040  & 0.019 & \underline{0.017} & 0.010 & \underline{0.015}  & 0.029 & 0.054 & 0.015 & 0.033\\
\cmidrule(l){1-13}
GraphVAE & 0.350 & 0.980 & 0.054 & 0.623  & 0.130 & 0.170 & 0.050 & 0.117  & 1.369 & 0.629 & 0.191 & 0.730 \\
GNF & 0.200 & 0.200 & 0.110 & 0.170  & 0.030 & 0.100 & \textbf{0.001} & 0.044 & - & - & - & - \\
GraphDF & 0.060 & 0.120 & 0.030 & 0.070  & 0.040 & 0.130 & 0.010 & 0.060  & 1.503 & 1.061 & 0.202 & 0.922\\
DGAE  & \underline{0.032} & 0.062 & \underline{0.005} & \underline{0.033}  & 0.021 & 0.041 & 0.007 & 0.023  & 0.020 & \underline{0.051} & 0.003 & \underline{0.025}\\
$\textbf{GLAD}$ & \textbf{0.029} & \underline{0.047} & 0.008 & \textbf{0.028}  & \textbf{0.012} & \textbf{0.013} & 0.004 & \textbf{0.010}  & \underline{0.012} & \textbf{0.014} & \textbf{0.001} & \textbf{0.009} \\
\bottomrule
\end{tabular}
\end{center}
\caption{Generation results on the generic graph datasets. We show the mean values of 15 runs for each experiment. The baselines are sourced from \cite{jo2022score, kong2023autoregressive}.  We compare with generative models operating on graph space (top row) and on latent-graph space (bottom row). Hyphen (-) denotes unreproducible results. The $1^\text{st}$ and $2^\text{nd}$ best results are bolded and underlined, respectively.}
\label{table:gen-data}
\end{table*}

\begin{table*}[t]
\begin{center}
\begin{tabular}{l ccccc ccccc}
\toprule
 & \multicolumn{5}{c}{\textsc{QM9}} & \multicolumn{5}{c}{\textsc{ZINC250k}}\\
\cmidrule(lr){2-6} \cmidrule(l){7-11}
 & Val. $\uparrow$ &  Uni. $\uparrow$ & Nov. $\uparrow$ & NSPDK $\downarrow$ & FCD $\downarrow$ &  Val. $\uparrow$ &  Uni. $\uparrow$ & Nov. $\uparrow$ & NSPDK $\downarrow$ & FCD $\downarrow$\\
\midrule
GraphAF & 74.43 & 88.64 & 86.59 & 0.020  & 5.27 & 68.47 & 98.64 & 100  & 0.044 & 16.02\\
MoFlow & 91.36 & 98.65 & 94.72 & 0.017 & 4.47 & 63.11 & 99.99 & 100 & 0.046 & 20.93\\
GDSS  & 95.72 & 98.46 & 86.27 & 0.003  & 2.90 & 97.01 & 99.64 & 100  & 0.019 & 14.66\\
DiGress & 99.0 & 96.66 & 33.40 & \underline{0.0005}  & \underline{0.360} & 91.02 & 81.23 & 100 & 0.082 & 23.06\\
GraphArm  & 90.25 & 95.62 & 70.39 & 0.002  & 1.22 & 88.23 & 99.46 & 100 & 0.055 & 16.26\\
\cmidrule(l){1-11}
GraphDF & 93.88 & 98.58 & 98.54 & 0.064  & 10.93 & 90.61 & 99.63 & 100  & 0.177 & 33.55\\
DGAE & 92.0 & 97.61 & 79.09 & 0.0015 & 0.86 & 77.9 & 99.94 & 99.97  & \underline{0.007} & \underline{4.4}\\
$\textbf{GLAD}$ & 97.12 & 97.52 & 38.75 & \textbf{0.0003} & \textbf{0.201} & 81.81 & 100 & 99.99 & \textbf{0.002} & \textbf{2.54}\\
\bottomrule
\end{tabular}
\end{center}
\caption{Generation results on the molecule datasets. We show the mean values of 3 runs for each experiment. The baselines are sourced from \cite{jo2022score, kong2023autoregressive}. We highlight the most important metrics; the $1^\text{st}$ and $2^\text{nd}$ best results are bolded and underlined, respectively. We compare with generative models operating on graph space (top row) and on latent-graph space (bottom row).}
\label{table:molecule}
\end{table*}

\section{Experiments}
\subsection{Graph Reconstruction}

\begin{figure}[t] 
\begin{center}
\centerline{\includegraphics[width=1.\columnwidth]{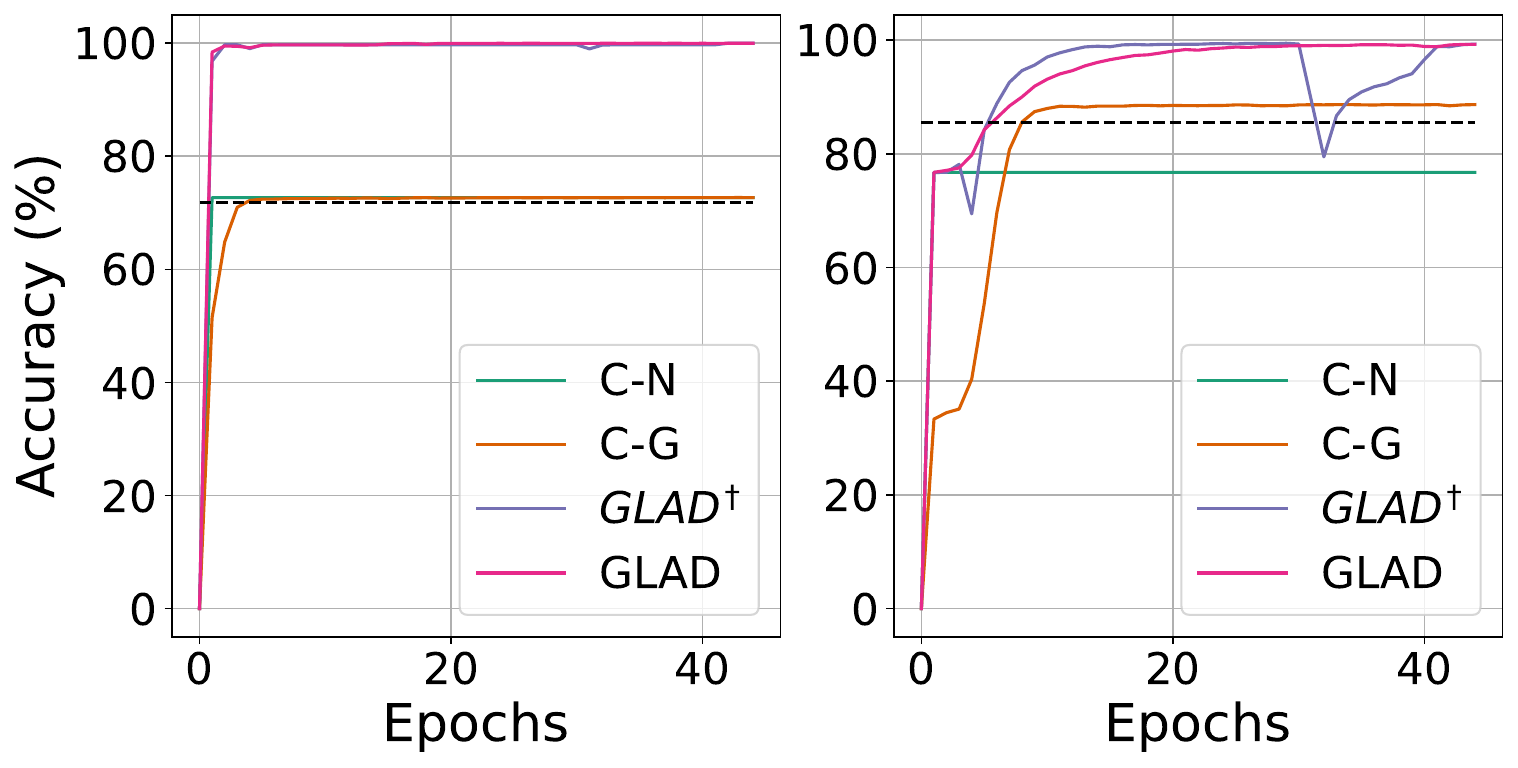}}
\caption{Molecule reconstruction from different latent spaces on QM9. Atom (Left) and bond (Right) type test accuracies: i) continuous-graph latent representation (C-G).
ii) continuous-node latent representation (C-N)  iii) unquantized discrete-node latent space $\text{GLAD}^{\dagger}$, iv) quantised discrete-node latent space GLAD. Dashed back lines denote the percentage of dominant atom (Carbon) and bond (Single) types.
}
\label{fig:latent-recon}
\end{center}
\end{figure}

\paragraph{Setup}
We claim that our  quantized-graph latent space is pivotal to the good performance of GLAD. We now provide empirical evidence to support our claim. We compare the reconstruction performance of GLAD applied on its quantized graph latent space to the following latent spaces and methods: i) continuous-graph latent space (C-G); we train a vanilla VAE on a global structure, pooled latent nodes, to get a continuous-latent representation ii) continuous-node latent space (C-N); similar to i) we also train a vanilla VAE, however applied on a set of latent nodes. Finally in iii) we simply use a set of raw node embeddings without any constraints imposed to the graph-latent space  ($\text{GLAD}^{\dagger}$), akin to a standard autoencoder for graphs. We  evaluate the quality of the underlying graph latent spaces using their bond-type (X) and atom-type (E) reconstruction accuracies on QM9.

\paragraph{Results}
We show the reconstruction ability per latent structure in Figure \ref{fig:latent-recon}. The two continuous latent spaces (C-G, C-N) have considerably lower accuracies compared to the two GLAD variants, which preserve to a different extent the original discrete graph topology. Both GLAD variants converge very fast to nearly $100\%$ reconstruction accuracy. GLAD has a small advantage when it comes to the bond type (E), for which its training is more stable than that of its unquantized sibling $\text{GLAD}^{\dagger}$. The latent spaces of  C-G and C-N suffer from the so-called latent-node-collapsing problem. They are not able to well describe different atom local structures. As a result, most reconstructed nodes are assigned to the dominant Carbon atom, roughly corresponding to the $72\%$ of atoms in QM9.

\subsection{Generic Graph Generation}

\paragraph{Setup} 
We measure GLAD's ability to capture the underlying structures of generic graphs on three datasets: (a) ego-small \cite{sen2008collective}, (b) community-small, and (c) enzymes \cite{schomburg2004brenda}. On these graphs, there is no explicit information about nodes available for training. We therefore use node degrees as augmented-node features to the graph-encoder's inputs. We use the same train- and test- splits as the baselines for a fair comparison. We evaluate model performance by computing the maximum-mean discrepancy (MMD) between the generated-set and test-set distributions of the following graph statistics: degree (Deg. $\downarrow$), clustering coefficient (Clus. $\downarrow$), and orbit 4-node occurrences (Orb. $\downarrow$). At each evaluation, we generate an equal number of graphs to the current test set. We adopt the Gaussian Earth Mover's Distance (EMD) kernel to calculate MMD thanks to its computational stability.

\paragraph{Baselines} 
Here under the approach working directly on graph space, we compare to auto-regressive  GraphRNN \cite{graphrnn}, auto-regressive flow GraphAF \cite{shi2020graphaf}, auto-regressive diffusion GraphArm \cite{kong2023autoregressive}, continuous diffusion GDSS \cite{jo2022score}, and discrete diffusion DiGress \cite{vignac2023digress}. Under the latent-space direction, we have GraphVAE \cite{simonovsky2018graphvae}, continuous flow GraphNF\cite{liu2019graph}, discrete flow GraphDF \cite{luo2021graphdf}, and discrete autoencoder DGAE \cite{boget2024discrete}. In all baselines, we use the continuous- and discrete- terms with an implicit indication on how data represented or treated by individual approach.

\paragraph{Results}
We observe in Table \ref{table:gen-data} that our model achieves competitive performance compared to the state-of-the-art baselines; it consistently maintains the lowest-average MMD distance across three datasets. Moreover, GLAD significantly outperforms all baselines that make explicit use of latent-space structures, namely GraphVAE, GNF, GraphDF, and DGAE.
These accomplishments would highlight two pivotal strengths of GLAD: first, our quantized graph-latent space can encode rich local graph sub-structures, serving as a cornerstone for any latent-driven generative models. Second, the diffusion bridges work seamlessly within the proposed latent structure by design, which is underscored by its capability to learn the set-based latent representation of graphs. While GLAD does not operate directly on the graph space like GDSS and DiGress, which are also diffusion-based frameworks, GLAD demonstrates superior performance to capture the underlying topologies of graphs in a holistic manner.

\subsection{Molecule Graph Generation}

\paragraph{Setup} 
We evaluate the capacity of GLAD to capture the complex dependencies between atoms and bonds as they appear in different molecules. We conduct experiments on two standard datasets: QM9 \cite{ramakrishnan2014quantum} and ZINC250k \cite{irwin2012zinc}. Following \cite{jo2022score}, we remove hydrogen atoms and kekulize molecules by RDKit \citet{landrum2016rdkit}. We quantitatively evaluate all methods in terms of validity without post-hoc corrections (Val. $\uparrow$), uniqueness (Uni. $\uparrow$), and novelty (Nov. $\uparrow$) of $10.000$ generated molecules. In addition, we compute two more salient metrics that quantify how well the distribution of the generated graphs aligns with the real data distribution, namely Fréchet ChemNet Distance (FCD $\downarrow$) and Neighborhood Subgraph Pairwise Distance Kernel (NSPDK $\downarrow$). FCD evaluates the distance in the chemical space between generated and training graphs by using the activation of the ChemNet's penultimate layer, while NSPDK measures the MMD distance, showing the similarity of underlying structures between generated and test molecules. These last two metrics show the most relevant comparisons as they directly take into account the distribution of molecular properties, e.g chemical and structural, rather than with only molecule-graph statistics. 

\paragraph{Baselines}
We compare GLAD with generative models that operate on graph spaces, including continuous flow GraphAF \cite{shi2020graphaf}, conditional flow MoFlow \cite{zang2020moflow}, continuous diffusion GDSS \cite{jo2022score}, discrete diffusion DiGress \cite{vignac2023digress}, and auto-regressive diffusion GraphArm \cite{kong2023autoregressive}. We also compare with generative models that work with discrete latent structures, including auto-regressive flow GraphDF \cite{luo2021graphdf}, and auto-regressive autoencoder DGAE \cite{boget2024discrete}.

\begin{figure}[t] 
\begin{center}
\centerline{\includegraphics[width=1.\columnwidth]{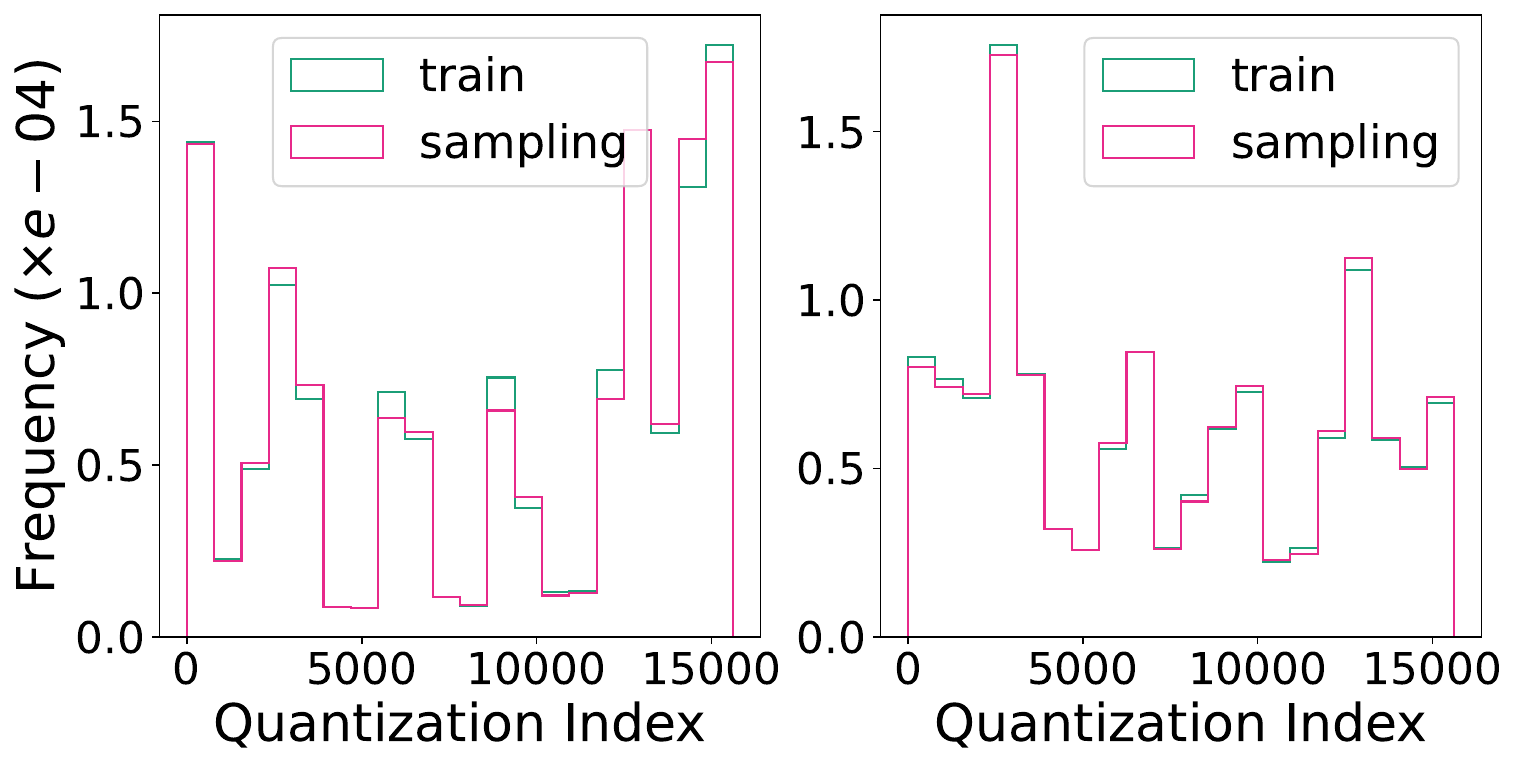}}
\caption{Comparison of latent node distributions on the quantized-grid structure $\mathbf{S}$ of train and sampled molecules, QM9 (Left) and ZINC250k (Right).}
\label{fig:latent-dist}
\end{center}
\end{figure}

\paragraph{Results} Table \ref{table:molecule} show the generation results on molecules. GLAD is the first one-shot model that learns to generate molecular structures from a discrete latent space. Even though our model does not learn directly on atom- and bond-type structures, GLAD  can still generate molecules with good validity scores without corrections. Importantly, GLAD achieves state of the art performance on the two most salient metrics NSPDK and FCD that capture well both structural and chemical properties of molecule distributions, consistently outperforming by significant margins all baselines. It demonstrates that our quantized latent space offers a suitable discrete structure to encode those properties. We additionally compare the latent node distributions between train and sampled molecules in Figure \ref{fig:latent-dist}, which shows the GLAD's bridge model is able of capturing well the latent node distributions of different datasets.

\subsection{Ablation Studies}

\begin{table}[t]
\begin{center}
\begin{small}
\begin{tabular}{l ccccc}
\toprule
\multirow{2}{*}{Prior}  & \multicolumn{5}{c}{\textsc{QM9}}\\
\cmidrule(l){2-6}
& Val. $\uparrow$ &  Uni. $\uparrow$ & Nov. $\uparrow$ & NSPDK $\downarrow$ & FCD $\downarrow$\\
\midrule
$\mathcal N( \mathbf{0}, \mathbf{1})$ & 95.38 & 97.38 & 41.54 & 0.0004 & 0.330\\
\midrule
$\mathbf 0$ & 97.12 & 97.52 & 38.75 & 0.0003 & 0.201\\
\midrule
$\mathbf 0^{\dagger}$ & 83.12 & 97.54 & 62.44 & 0.0009 & 1.207 \\
\bottomrule
\end{tabular}
\end{small}
\end{center}
\caption{Ablations on prior $\mathbb{P}_0$ and quantization $\mathcal{F}$. Generation results when the prior distribution is initialized as a standard normal distribution and a Dirac $\Delta$ distribution on the fixed point $\mathbf 0$; the latter is ablated with and without quantization, denoted by $\mathbf 0$ and $\mathbf 0^{\dagger}$ respectively.}
\label{tab:prior}
\end{table}

\paragraph{Quantisation effect on generation}
We study the effect of quantisation by removing the quantization step. Node embeddings remain discrete-, but non-quantised- structures, and they are embedded into a continuous domain, which has the same dimensionality to the quantized one $f$, denoted as $\Omega_c = [L_{\min}, L_{\max}]^{(N \times f)}$. We retrain our graph autoencoder, drift model , and obtain $L_{\min}, L_{\max}$ that correspond to the min- and max- values of learned latent nodes for the entire training set. The $\Pi$-bridge's drift on a continuous domain is computed as:
\begin{align}
    \left[\eta^{\Pi} \right]^{h} &= \sigma_t^2 \nabla_{Z_t^h} \log \left(F(\frac{Z_t^h-L_{\min}}{\sqrt{\beta_T - \beta_t}}) - F(\frac{Z_t^h-L_{\max}}{\sqrt{\beta_T - \beta_t}}) \right) \nonumber\\ & \forall h,  1 \leqslant h  \leqslant (N \times f)
\end{align}
Where $Z_t \in [L_{\min}, L_{\max}]^{(N \times f)}$, $F$ is the standard Gaussian Cumulative Distribution Function (CDF).
As per Figure \ref{fig:latent-recon}, the two GLAD variants have a rather similar reconstruction performance, but this is not the case when we compare their generation performance, described in Table \ref{tab:prior}. There we see that the diffusion bridges can not capture well the graph distribution when they operate on the unquantized space, with a performance drop for most measures (importantly on the two measures capturing molecular distributions, NSPDK and FCD), showing the benefits of the quantised discrete latent space. Quantization acts as a spatial regulariser over the non-quantised structures by constraining latent nodes on high-dimensional discretized grid instead of letting them localized in a infinite-continuous space. We hypothesize this spatial regularisation is non-trivial to construct effective latent structures for graphs.

\paragraph{Choice of bridge priors}
Here we evaluate model performance on different priors;  a fixed point prior $\mathbb{P}_0 = \mathbf{0}$, which is the central mass of our quantized latent space, and a standard normal distribution prior, $\mathbb{P}_0 = \mathcal{N}( \mathbf {0}, \mathbf {I})$. We conduct the experiments on QM9 and give the results in Table \ref{tab:prior}.
As a result of the steering forces of the bridge processes, there are only negligible differences between the two priors making prior selection less of a nuance. With the fixed prior, we obtain generated molecules with higher validity scores, closer to the training distribution in both chemical space (FCD score), structural space (NSPDK score). Using the normal prior we have slightly better performance in terms of novelty score.

\section{Conclusion}
We present the first equivariant graph diffusion model that operates over a discrete latent space. By the careful design, our graph-latent structure satisfies certain desiderata that stem from the graph nature, namely it should allow for learning graph-permutation-invariant distributions, being rich representational power, and respecting the inherently discrete nature of graphs. We empirically validate GLAD on a number of benchmarks and show that it achieves state of the art generative performance, surpassing other baselines independently of whether they operate on the original or on latent spaces. We systematically ablated different design choices for graph latent space representation and show that our discrete latent space brings clear performance advantages over continuous alternatives both in terms of reconstruction as well as generation. This paves the way for a more systematic exploration of graph generative modelling in latent spaces, something that until now has received rather limited attention.

\section{Acknowledgements}
 We acknowledge the financial support of the Swiss National Science Foundation within the LegoMol project (grant no. 207428). The computations were performed at the University of Geneva on Baobab and Yggdrasil HPC clusters.
 
\bibliography{aaai25}

\newpage
\onecolumn

\section{Technical Appendix}
\subsection{Discrete Latent Graph Diffusion Derivations}
We adjust the derivations of \citet{liu2023learning} to learning on the graph setting. We start by considering Brownian motion as the non-conditional diffusion process:
\begin{equation}
\label{eq:brownian}
    \mathbb{Q}: \;\; dZ_t = \sigma_t dW_t, \;\;  \;\; \beta_t = \int_0^t\sigma_s^2ds \nonumber
\end{equation}
The transition probability between states follows a Gaussian distribution, which density is defined as $q_{T|t} (Z_T | Z_t)=\mathcal{N}\left(Z_T; Z_t, \beta_T - \beta_t\right)$. Applying this density, we get the $Z_{\mathcal{G}}$-bridge drift function:
\begin{align}
    \eta^{Z_{\mathcal{G}}}(Z_t,t) &= \sigma^2_t\nabla_{Z_t}\log q_{T|t}(Z_{\mathcal{G}}|Z_t) \nonumber\\ 
                  &=\sigma^2_t\frac{Z_{\mathcal{G}}- Z_t}{\beta_T -\beta_t}
\end{align}
Thus, we can derive the governing law of the $Z_{\mathcal{G}}$-bridge:
\begin{align}
    \mathbb{Q}^{Z_{\mathcal{G}}}: \;\; dZ_t = \sigma_t^2 \frac{Z_{\mathcal{G}} - Z_t}{\beta_T - \beta_t}dt + \sigma_tdW_t \nonumber
\end{align}
And we sample from the $t \in [0,T]$ time step of the $Z_{\mathcal{G}}$-bridge as follows:
\begin{align}
    Z_t =  \frac{\beta_t}{\beta_T}\left(Z_{\mathcal{G}} + (\beta_T - \beta_t)Z_0\right)+ \xi\sqrt{\beta_t\left(1-\frac{\beta_t}{\beta_T}\right)},\;\;  \xi \sim \mathcal{N}(0, I),\;\;  Z_0 \sim \mathbb{P}_0 \nonumber
\end{align}
Where $\mathbb{P}_0$ is the bridge prior, we ablate two cases: a fixed prior $\mathbb{P}_0 := \mathbf{0}$, and a standard Gaussian prior $\mathbb{P}_0 := \mathcal{N}(0, I)$.
By plugging the probability $q_{T|t} (Z_T | Z_t)$, we obtain the $\Pi$-bridge drift:
\begin{align}
\eta^{\Pi}(Z_t,t) &= \sigma^2(Z_t,t)\mathbb{E}_{\omega \sim q_{T|t,\Omega}(\omega|Z_t)}[\nabla_{Z_t}\log q_{T|t}(\omega|Z_t)] \nonumber\\ 
                  &= \sigma^2_t\mathbb{E}_{\omega \sim q_{T|t,\Omega}(\omega|Z_t)}\left[\frac{\omega- Z_t}{\beta_T -\beta_t}\right] \nonumber
\end{align}
with $Z_t$ samples from the $Z_{\mathcal{G}}$-bridge, $\omega$ is sampled from a transition density given by $q_{T|t, \Omega}(\omega|Z_t) := q(Z_T = \omega| Z_t, Z_T \in \Omega)$. While $\mathbb{Q}$ is Brownian motion, the transition probability can be truncated into a mixture of Gaussian densities over each element $\omega \in \Omega$, which is given by  $q_{T|t,\Omega}(\omega|Z_t)  \varpropto \mathbb{I}(\omega \in \Omega) q_{T|t}(Z_T = \omega| Z_t)$.
We observe that the truncated transition density considers the existing elements $\omega$ of the latent-graph domain $\Omega$. However, our graph latent representation is a set-based structure, the discrete domain $\Omega$ can thus be factorized over latent node dimension, denoted as $\Omega = I_1 \times \cdots \times I_n$, with $I_i = \mathbf{S}$. Intuitively, this factorization represents for independent conditional stochastic processes over each latent node dimension of the given graph latent structure. Hence, the $\Pi$- bridge's drift can be decomposed as:
\begin{align}
    & \eta^{\Pi}(Z_t,t) = \left[ \eta^{I_i}(Z_t^i, t) \right]_{i=1}^N \nonumber\\
    & \text{where }\eta^{I_i}(Z_t^i, t) = \sigma^2_t\mathbb{E}_{s_k \sim q_{T|t,\mathbf{S}}(s_k|Z_t^i)}\left[\frac{s_k- Z_t^i}{\beta_T -\beta_t}\right] \nonumber
\end{align}

The expectation is now taken over the truncated transition probability along each latent node dimension $I_i$, whose density can be represented as following:

\begin{align}
    & q_{T|t,\textbf{S}}(s_k|Z_t^i) \varpropto \mathbb{I}(s_k \in \textbf{S})\frac{q_{T|t}(s_k|Z_t^i)}{\sum_{j=1}^K q_{T|t}(s_j|Z_t^i)} \nonumber\\
    & \text{where the conditional density } q_{T|t}(s_j|Z_t^i) \text{ is } \mathcal{N}(s_j;Z_t^i, \beta_T - \beta_t) \nonumber
\end{align}

The $\Pi$-bridge's drift on latent node $I_i$  can be further developed as:
\begin{align}
    \eta^{I_i}(Z_t^i, t) &= \sigma^2_t\mathbb{E}_{s_k \sim q_{T|t,\mathbf{S}}(s_k|Z_t^i)}\left[\frac{s_k- Z_t^i}{\beta_T -\beta_t}\right] \nonumber\\
         &= \sigma^2_t \sum_{k=1}^K \frac{q_{T|t}(s_k|Z_t^i)}{\sum_{\sum_{j=1}^K q_{T|t}(s_j|Z_t^i)}}\frac{s_k- Z_t^i}{\beta_T -\beta_t} \nonumber\\
         &= \sigma^2_t \sum_{k=1}^K \frac{\exp\left(-\frac{\|Z_t^i - s_k\|^2}{2\left(\beta_T - \beta_t\right)}\right)}{\sum_{j=1}^K \exp\left(-\frac{\|Z_t^i - s_j\|^2}{2\left(\beta_T - \beta_t\right)}\right)}\frac{s_k- Z_t^i}{\beta_T -\beta_t}\nonumber\\
         &= \sigma_t^2 \nabla_{Z_t^{I_i}}\log\sum_{k=1}^K \exp\left(-\frac{\| Z_t^{I_i} - s_k \|^2}{2(\beta_T - \beta_t)}\right)
\end{align}

Following \cite{liu2023learning} we train our latent-graph model bridge $\mathbb{P}^{\psi}$ by  minimizing the Kullback-Leibler divergence between the probability-path measures of the model bridge  and the $\Pi$-bridge, $\min_\psi \{ \mathcal{KL}(\mathbb{Q}^{\Pi} || \mathbb{P}^{\psi})\}$, which is equivalent to the minimization of expectation over $\mathcal{KL}\left(\mathbb{Q}^{Z_{\mathcal{G}}} \| \mathbb{P}^{\psi} \right)$ up to an additive constant:

\begin{align}
    \mathcal{L} = \mathcal{KL}(\mathbb{Q}^{\Pi} || \mathbb{P}^{\psi}) = \mathbb{E}_{Z_{\mathcal{G}} \sim \Pi} \left[\mathcal{KL}\left(\mathbb{Q}^{Z_{\mathcal{G}}} \| \mathbb{P}^{\psi} \right)\right] + const. \nonumber
\end{align}

By Girsanov theorem \cite{lejay2018girsanov}, the $\mathcal{KL}$ divergence between the probability-path measures of the model-bridge and the $Z_{\mathcal{G}}$-bridge can be further decomposed:

\begin{align}
    \label{eq:raw_obj}
    \mathcal{L}= \mathbb{E}_{Z_{\mathcal{G}} \sim \Pi, Z_t \sim \mathbb{Q}^{Z_{\mathcal{G}}}} \left[-\log p_0(Z_0) + \frac{1}{2} \int_0^T \left\| \sigma_t^{-1}(\eta^{\psi}(Z_t, t) - \eta^{Z_{\mathcal{G}}}) \right\|^2\right] + const. 
\end{align}

Where $\eta^{\psi}(Z_t, t)$ is the parametric model-bridge's drift with a neural network. The prior distribution $p_0(Z_0)$ is a Dirac $\Delta$ distribution on the fixed point $\mathbf 0$, the central mass of $\mathbf{S}$, thus $\log p_0(Z_0)$ is constant. Applying the definition of $\eta^{\psi}(Z_t, t)$ to Equation \ref{eq:raw_obj}, the objective can be obtained with a closed-form solution:

\begin{align}
    \mathcal{L}(\psi) = \mathbb{E}_{t \sim [0,T], Z_{\mathcal{G}} \sim \Pi, Z_t \sim \mathbb{Q}^{Z_{\mathcal{G}}}} \left[\left\|{\psi}(Z_t,t) - \sigma^{-1}_t(\eta^{Z_{\mathcal{G}}}(Z_t,t) - \eta^{\Pi}(Z_t,t))\right\|^2\right] + const
\end{align}

In sampling new latent-graph samples, we apply the Euler–Maruyama discretization on the model bridge $\mathbb{P}^{\psi}$ with a  discretization step $\delta_t$:

\begin{align}
    Z_{t+1} = Z_{t} + \left(\sigma_t{\psi}(Z_t,t) + \eta^{\Pi}(Z_t,t)\right)\delta_t + \sigma_t\sqrt{\delta_t}\xi, \;\; \xi \sim \mathcal{N}(0, I),  \;\; Z_0 \sim \mathbb{P}_{0}
\end{align}

\subsection{Finite Scalar Quantization for Graphs}
We empirically find that the quantized latent space of dimension $f = 6$ works well to encode graphs within low reconstruction errors. In all experiments, we fix the quantization vector $L=[5,5,5,5,5,5]$ where at each latent node dimension $L_j,\; 1  \leqslant j  \leqslant 6$, we quantize to one of five possible values $[-2,-1,0,1,2]$. In total, we have $K=5^6$ quantization points that are uniformly distributed in the discrete latent space. These points serve as the referent points to which latent nodes are assigned. For illustration, given a raw latent node vector $Z^i$, its quantized version $Z^i_q$ is obtained as following:

\begin{align*}
    Z^i = \left[  \begin{array}{c}-1.1 \\ -1.7 \\ -0.01 \\ 0.1 \\ 3.2 \\ 0.6 \end{array} \right] \longrightarrow \frac{L_j}{2}\tanh{Z^i_j}  \longrightarrow \left[\begin{array}{c} -2.00 \\ -2.34 \\ -0.03 \\  0.25 \\  2.49 \\  1.34 \end{array} \right] \longrightarrow \text{Rounding} \longrightarrow Z^i_q = \left[  \begin{array}{c} -2 \\ -2 \\ 0 \\ 0 \\ 2 \\ 1 \end{array} \right]
\end{align*}

The same principle is applied to the other latent nodes.

\subsection{Experimental Details}

\paragraph{Hardware} We train GLAD on HPC system where we set 8 CPU cores and TITAN RTX for all experiments. For the generic graph datasets, we use one GPU  for both training and sampling. For molecule graphs, we train on 4 GPUs to accelerate training with large batch sizes. During inference, we use one GPU for QM9 and two GPUs for ZINC.

\paragraph{Hyperparameters} We conduct an ablation study on architecture sizes with different number of layers $\{4, 6, 8\}$, hidden node features $\{64, 128, 256\}$, and hidden edge features $\{32, 64, 128\}$. We resume GLAD's hyperparameter in Table \ref{table:glad-hyper}.

\paragraph{Data statistics} 
We validate GLAD on two types of graphs that include generic graphs and molecules. We additionally provide these graph statistics in Table \ref{table:stat-gen} and \ref{table:molecule_stats}.

\begin{table*}[ht]
\caption{\textbf{Generic graph statistics} on community small, ego small, and enzymes.}
\label{table:stat-gen}
\begin{center}
\begin{small}
\begin{tabular}{lcccc}
\toprule
 & Number of Graphs &  Number of Nodes &  Real / Synthetic \\
\midrule
\textsc{Community-small} & $100$ & $12  \leqslant |X| \leqslant 20$ & Synthetic\\
\textsc{Ego-small} & $200$ & $4  \leqslant |X| \leqslant 18$  & Real\\
\textsc{Enzymes} & $587$ &  $10  \leqslant |X| \leqslant 125$ & Real\\
\bottomrule
\end{tabular}
\end{small}
\end{center}
\end{table*}

\begin{table*}[ht]
\caption{\textbf{Molecule graph statistics} on QM9 and ZINC250k.}
\label{table:molecule_stats}
\begin{center}
\begin{small}
\begin{tabular}{lcccc}
\toprule
 & Number of Graphs &  Number of Nodes &  Number of Node Types & Number of Edge Types \\
\midrule
\textsc{QM9} & $133885$ & $1  \leqslant |X| \leqslant 9$ & $4$ & $3$ \\
\textsc{ZINC250k} & $249455$ & $6  \leqslant |X| \leqslant 38$  & $9$ & $3$\\
\bottomrule
\end{tabular}
\end{small}
\end{center}
\end{table*}

\paragraph{QM9's novelty issue} The dataset comprises all enumerated small molecules that compose from only four atoms C, N, O, F. Hence, the generation of novel graphs beyond this dataset might not accurately reflect the model's ability to capture the underlying data distribution. Typically, models that capture better the chemical property (FCD) distribution of QM9 tend to exhibit lower novelty scores. However, such problem can be alleviated when we train models on large-scale datasets such as ZINC250k.

\begin{figure*}[ht]
\begin{center}
\centerline{\includegraphics[width=13cm]{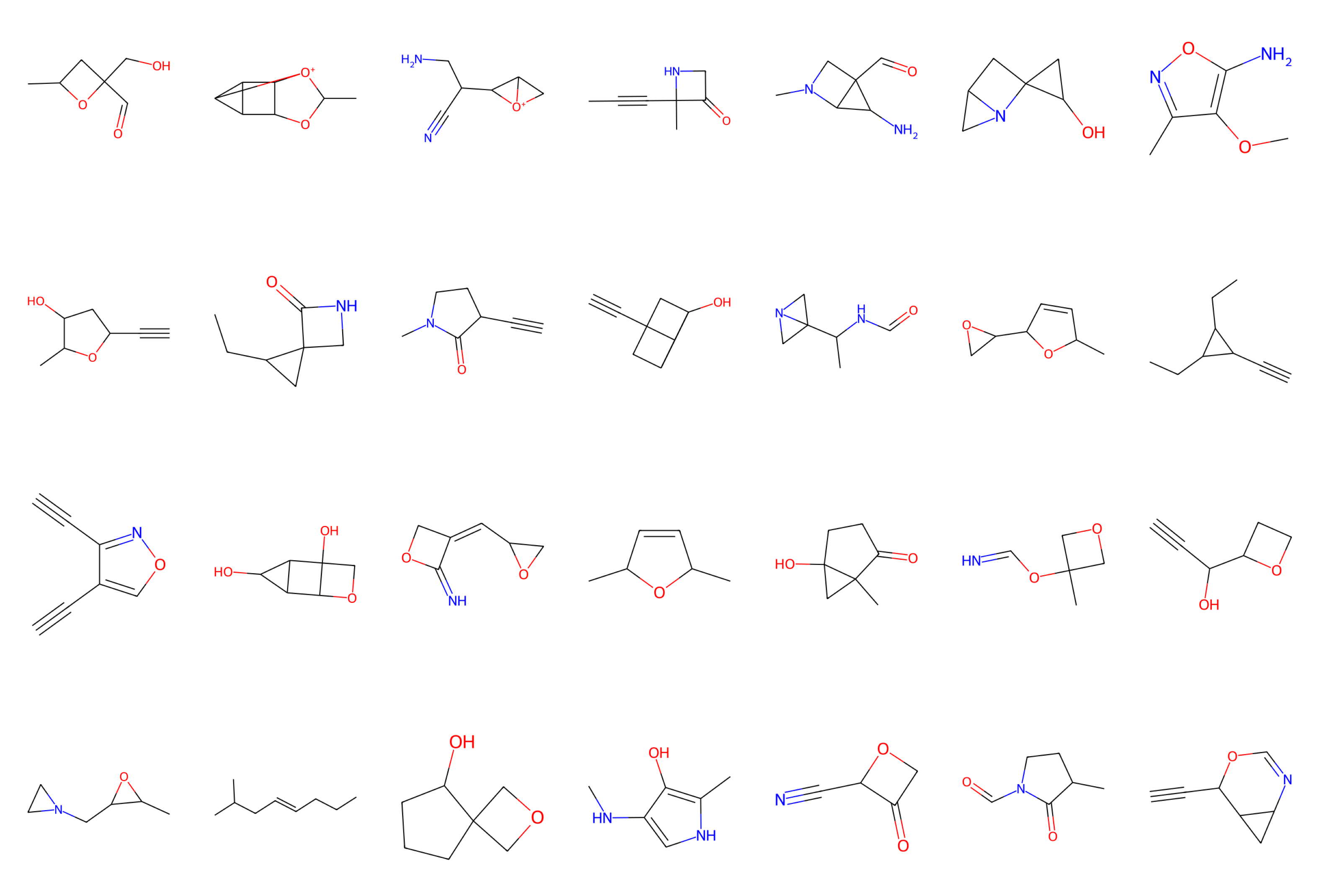}}
\caption{\textbf{Visualization of generated samples on QM9 from GLAD.}}
\label{fig:qm9_gen}
\end{center}
\end{figure*}

\begin{figure*}[t]
\begin{center}
\centerline{\includegraphics[width=15cm]{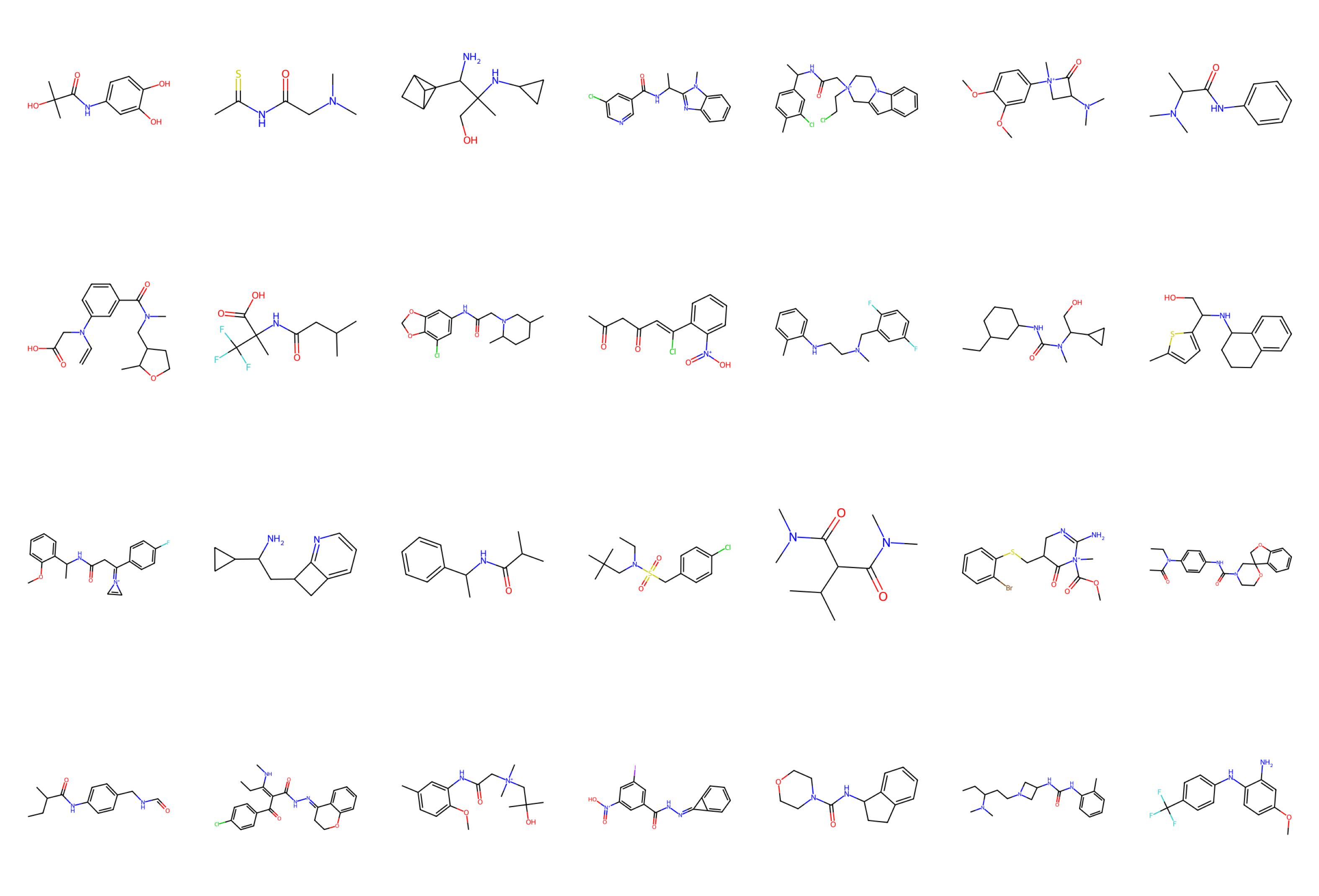}}
\caption{\textbf{Visualization of generated samples on ZINC250k from GLAD.}}
\label{fig:zinc250k_gen}
\end{center}
\end{figure*}

\paragraph{Detailed results} We show the detailed results by means and standard deviations for each experiment in Table \ref{table:gen-detail} and \ref{table:mol-detail}. We provide the visualizations of generated molecules from GLAD in Figure \ref{fig:qm9_gen} and \ref{fig:zinc250k_gen}.

\begin{table*}[ht]
\caption{\textbf{Detailed results on generic graphs}. We show the means and standard deviations of 15 runs.}
\label{table:gen-detail}
\vskip 0.15in
\begin{center}
\begin{small}
\begin{tabular}{lccc} 
\toprule
 & Deg. $\downarrow$ &  Clus. $\downarrow$ & Orb. $\downarrow$ \\ 
\midrule
\textsc{Community-small} & 0.029 $\pm$ 0.020 & 0.047 $\pm$ 0.025 & 0.008 $\pm$ 0.006 \\ 
\textsc{Ego-small} & 0.012 $\pm$ 0.007 & 0.013 $\pm$ 0.006 & 0.004 $\pm$ 0.002 \\  
\textsc{Enzymes} & 0.012 $\pm$ 0.004 & 0.014 $\pm$ 0.003 & 0.001 $\pm$ 0.000 \\ 
\bottomrule
\end{tabular}
\end{small}
\end{center}
\vskip -0.1in
\end{table*}

\begin{table*}[ht]
\caption{\textbf{Detailed results on molecule graphs, bridge-prior- and quantisation- ablations}. We show the means and standard deviations of 3 runs. $\text{ZINC250k}$ and $\text{QM9}$ are GLAD, i.e. fixed prior and quantised discrete latent space. 
$\text{QM9}^{\star}$ ablates the use of standard normal distribution for the bridge prior.  
$\text{QM9}^{\dagger}$ ablates the effect of non-quantisation in the latent space. We also report the validity of generated molecules with correction by RDKit.}

\label{table:mol-detail}
\vskip 0.15in
\begin{center}
\begin{small}
\begin{tabular}{lcccccc} 
\toprule
 & Validity with correction $\uparrow$ & Val. $\uparrow$ &  Uni. $\uparrow$ & Nov. $\uparrow$ & NSPDK $\downarrow$ & FCD $\downarrow$ \\ 
\midrule
\textsc{$\text{ZINC250k}$} & 100 $\pm$ 0.00 & 81.81 $\pm$ 0.29 & 100 $\pm$ 0.00 & 99.99 $\pm$ 0.01 & 0.0021 $\pm$ 0.0001 & 2.54 $\pm$ 0.05 \\ 
\textsc{$\text{QM9}$}& 100 $\pm$ 0.00 & 97.12 $\pm$ 0.03 & 97.52 $\pm$ 0.06 & 38.75 $\pm$ 0.25 & 0.0003 $\pm$ 0.0000 & 0.201 $\pm$ 0.003 \\ 
\textsc{$\text{QM9}^{\star}$}& 100 $\pm$ 0.00 & 95.38 $\pm$ 0.00 & 97.38 $\pm$ 0.00 & 41.54$\pm$ 0.01 & 0.0004 $\pm$ 0.0000 & 0.330 $\pm$ 0.013 \\ 
\textsc{$\text{QM9}^{\dagger}$}& 100 $\pm$ 0.00 & 83.12 $\pm$ 0.01 & 97.54 $\pm$ 0.01 & 62.44 $\pm$ 0.53 & 0.0009 $\pm$ 0.0000 & 1.207 $\pm$ 0.003 \\ 
\bottomrule
\end{tabular}
\end{small}
\end{center}
\vskip -0.1in
\end{table*}

\begin{table*}[b]
\begin{center}
\begin{small}
\begin{tabular}{ll cccccc}
\toprule
 & Hyperparameter & Community-small & Ego-small & Enzymes & QM9 & ZINC250k \\
\midrule
\multirow{5}{*}{$\phi$} &  Number of heads & 8 & 8 & 8 & 8 & 8 \\
 & Number of layers & 8 & 8 & 8 & 8 & 8 \\
 & Hidden dimension X & 256 & 256 & 128 & 256 & 256 \\
 & Hidden dimension E & 128 & 128 & 32 & 128 & 128 \\
 & Hidden dimension Y & 64 & 64 & 64 & 64 & 64 \\
\midrule
\multirow{5}{*}{$\theta$} &  Number of heads & 8 & 8 & 8 & 8 & 8 \\
 & Number of layers & 4 & 4 & 4 & 4 & 4 \\
 & Hidden dimension X & 256 & 256 & 128 & 256 & 256 \\
 & Hidden dimension E & 128 & 128 & 32 & 128 & 128 \\
 & Hidden dimension Y & 64 & 64 & 64 & 64 & 64 \\
\midrule
\multirow{5}{*}{$f_\psi$} &  Number of heads & 8 & 8 & 8 & 8 & 8 \\
 & Number of layers & 4 & 4 & 8 & 8 & 8 \\
 & Hidden dimension X & 256 & 256 & 256 & 256 & 256\\
 & Hidden dimension E & 128 & 128 & 128 & 128 & 128 \\
 & Hidden dimension Y & 64 & 64 & 64 & 64 & 64 \\
\midrule
\multirow{3}{*}{SDE} & $\sigma_{\min}$ & 1.0 & 1.0 & 1.0 & 1.0& 1.0\\
 & $\sigma_{\max}$ & 3.0 & 3.0 & 3.0 & 3.0 & 3.0 \\
 & Number of diffusion steps & 1000 & 1000 & 1000 & 1000 & 1000 \\
\midrule
\multirow{4}{*}{AE} & Optimizer  & Adam & Adam & Adam & Adam & Adam \\
& Learning rate  & $5e^{-4}$ & $5e^{-4}$ & $5e^{-4}$ & $5e^{-4}$ & $5e^{-4}$ \\
& Batch size  & 32 &32 & 32 & 5240 & 512 \\
& Number of epochs  & 3000 & 3000 & 3000 & 1000 & 1000 \\
\midrule
\multirow{5}{*}{Model Bridge} & Optimizer  & Adam  & Adam & Adam & Adam & Adam \\
& Learning rate & $[1e^{-4}, 5e^{-4}]$ & $[1e^{-4}, 5e^{-4}]$ & $[1e^{-4}, 5e^{-4}]$ & $[1e^{-4}, 2e^{-3}]$ & $[3e^{-4}, 2e^{-3}]$ \\
& Batch size  & 64 & 64 & 64 & 1280 &  512 \\
& Number of epochs  & 5000 & 5000 & 5000 & 2000 & 2000 \\
\bottomrule
\end{tabular}
\end{small}
\end{center}
\caption{\textbf{GLAD hyperparameters }. We show the main hyperparameters for the generic and molecule generation tasks. We provide those hyperparameters to encoder $\phi$, decoder $\theta$, drift model $\psi$, graph autoencoder (AE), model bridge (Model Bridge), and diffusion process (SDE).}
\label{table:glad-hyper}
\end{table*}

\end{document}